\documentclass[letterpaper]{article}
\usepackage[preprint]{aaai2027}
\usepackage[hyphens]{url}
\usepackage{graphicx}
\usepackage{natbib}
\usepackage{caption}
\usepackage{booktabs}
\usepackage{array}

\renewcommand{\textfraction}{0.08}
\title{The More the Better?\\A Comparison of Single- and Multi-Agent LLM Setups on a Financial Review Task}
\author{Pei Liu}
\affiliations{
    School of Engineering and Applied Science, George Washington University\\
    Washington, DC, USA\\
    pei.liu@gwu.edu\\
    ORCID \url{https://orcid.org/0009-0001-9092-1086}
}
\begin{document}
\maketitle
\begin{abstract}
Practitioner guidance holds that a large language model reviews a document better when the work is divided among agents. We test a narrow form of it in a controlled comparison of single-agent and multi-agent setups on a financial task, reviewing credit assessment memos against requirements. One agent checks six requirements in one call. A team of three on the same model checks the same memo, two requirements each, in the same words, with no roles or later stage. Labels are known for every requirement on 300 synthetic memos, so misses and false alarms are measured separately. Two primary predictions are tested on four models at a Bonferroni-corrected level. Under an instruction to decide each requirement once, the team catches 0.08 to 0.10 more of the defects that carry an explanation on the three models whose single agent misses some, at the corrected level on two. It raises more false alarms on all four, at the corrected level on three. Balanced accuracy, an exploratory measure, changes by 0.02 or less, and the team uses 2.2 to 2.4 times the tokens. Showing each agent the requirements it was not given reduces the added false alarms on the two Claude models, at the corrected level on one. In exploratory comparisons, sampling the single agent three times and flagging on any draw gives higher balanced accuracy than the team on those models, and the team's results depend on the instruction.

\end{abstract}

\section{Introduction}

Checking a document against a list of written requirements is routine work in finance, where a credit file, an audit paper or a disclosure must satisfy each one. A common way to build such a review with a large language model is to split it among agents. One agent checks the figures, another the calculations, a third the judgement calls, and their findings are merged. A practitioner guide states the idea as a rule, that ``for complex tasks with multiple considerations, LLMs generally perform better when each consideration is handled by a separate LLM call'' \citep{schluntz2024building}. The guide offers no data in support.

Existing tests of the rule leave three things out. Among those that divide the work without adding stages, none reports a reviewer's miss rate together with a false-alarm rate measured on cases known to be free of defects. None gives the single agent as many calls as the divided reviewer, and none separates dividing the work from withholding what the other parts cover. A team usually differs from the single agent in more than the division of work, through role descriptions, an orchestrator, a filter or a different model for one of its agents.

A designer needs to know more than which arrangement scores higher. A review that catches more defects and raises more false alarms is a different instrument, and whether it is a better one depends on what each error costs.

We compare two arrangements with only the allocation of requirements to calls varied. A single agent receives six requirements in one call. A team of three agents on the same model receives the same six in identical words, two to a call, with no roles and no later stage. The documents are synthetic credit memos whose labels are known by construction. The comparison is a narrow form of the guide's rule. We give two considerations to a call where the rule speaks of one. Our task is six checks of one kind on a short document, where the rule speaks of complex tasks. Our main condition also switches off the models' separate thinking channel and tells the reviewer to decide each requirement once.

Two further arms separate the effect of division from the effects of what accompanies it. One samples the single agent's prompt three times, so that it uses as many calls as the team. The other, added after the first results had been read, shows each team agent the requirements it was not given.

Our contribution is to supply all three missing elements in one design. The trade we find, more defects caught for more false alarms, reproduces earlier work. The size of its false-alarm side, measured on cells free of defects, is new, and so is the contrast with a team whose agents see what the others cover. Two exploratory results are also new, a comparison of the trade with repeated draws of the single agent and its dependence on the instruction.

\section{Related work}

\textbf{Teams against single agents.} \citet{tran2026single} give a single agent and five multi-agent designs the same cap on thinking tokens and find the single agent ahead on average. \citet{zhang2025stop} find that no debate framework beats chain-of-thought prompting in more than a fifth of their settings. \citet{kim2025scaling} relate a team's benefit to task properties and to the single agent's own score. These studies score answers as right or wrong and so cannot say which of a reviewer's two errors a design trades for the other.

\textbf{Division of a detection task.} On financial documents, \citet{he2026fined} ask a model either for all error types at once or for one type in each query. With the whole document in the prompt, asking for one type at a time raises average recall by 0.08 to 0.10 and cuts precision by more than half [their Table 6]. The pattern reverses when the document is supplied in chunks, and they conclude from F1 that asking for all types at once is better. \citet{liu2026dude} find that a team with no filtering stage, on one model, raises recall from 58.7 to 75.0 percent and lowers precision from 91.5 to 74.2 [their Table 2], and that a single agent given the team's instruction also gains recall and loses precision, by a smaller amount. Their filter restores the precision. With added stages or debate a team can instead raise fewer false alarms and catch less. \citet{li2026optargus} report a team with a conductor, a judge and a reranker that passes 85 percent of defect-free artifacts against 48 for a single detector. With the reranker removed the team still passes 64 percent, and they test no team with all of those stages removed. \citet{chen2025debate} find that two earlier debate protocols mostly raise precision and lower recall, and \citet{darcy2024marg} that a team given the single agent's task prompt matches fewer of the human reviewers' comments at slightly higher precision.

\textbf{Criteria in one call or several.} Two studies vary only the criteria in a judge call, as we do. \citet{peng2026ruverbench} find that balanced accuracy falls once a call holds four or five rubrics on long coding-agent trajectories, and changes little on shorter reports. \citet{rao2026jev} find that on their two binary checklists a whole rubric in one call and one criterion at a time are equally accurate, while 3 to 14 percent of individual verdicts differ. Neither reports the miss rate and the false-alarm rate for each arrangement. \citet{yu2026bottleneck} attribute a team's benefit to isolated contexts and find that a single agent given the team's divided instructions in one context scores below the single agent with undivided instructions on most tasks [their Table 3]. Work on judges that assess the quality of an output, as opposed to detecting defects in a document, gives mixed results on the same question \citep{lee2025checkeval,xiong2025multicrit,sunkavalli2026raters}, and none of those tasks has defect-free cases.

\textbf{Repeated calls.} A fair baseline lets the single agent make as many calls as the team. Sampling a model several times and taking the most frequent answer goes back to self-consistency decoding \citep{wang2023self}. \citet{li2024more} treat each sample of one model as an agent and report that accuracy under a vote rises with the number of samples, up to 40, on reasoning and code benchmarks. \citet{chen2024more} show that a majority over repeated calls raises accuracy on queries where the correct answer is the single call's most likely answer and lowers it on the others. \citet{jain2025consensus} let four of fourteen judge models reject an item. \citet{teo2026beyond} count a contradiction as detected when any of several reviews finds it, and report no false-alarm rate for the reviewing system.

\textbf{Framing and gap.} We take our framing from \citet{fox2026omission}. Of every remedy applied to a judge they ask whether it ``slides the point along a curve or lifts the curve'', and they find that wording, voting and prompt search slide it. What lifts it in their study is a two-stage pipeline that extracts the facts to check and verifies each in its own call. The pipeline adds a stage and a closed output format, so it is not a division of one task among calls of the kind we test. Two controls in the studies above resemble the scoped team that Section 3.3 defines, and differ from it. Liu et al. give a single agent the team's instruction, and Yu et al. give a single agent the team's divided instructions in one context. Both alter the single agent. The scoped team leaves the division in place and restores information to the divided agents. Among the studies we examined, we found none that sets a division without added stages against repeated draws, or that separates division from the loss of that information.

\section{Experimental design}

\subsection{Setting}

We study one situation. Several agents built on the same model each receive the same complete document, and each is responsible for part of a single review whose parts draw on overlapping passages. The agents run in parallel, do not communicate and have no tools, and the document fits comfortably in one context window. A team can help for reasons that this design excludes, such as a task that does not fit in one context, subtasks that share no input, or agents with different models or tools. Our result says nothing about those cases.

\subsection{Task and ground truth}

The task is the review of a credit assessment memo against six written requirements. Each memo is written by a program, with no language model, from one record of a public credit card dataset \citep{yeh2009comparisons,yeh2009dataset}. A memo has about 1,100 words in eight sections, and we use 300 of them.

\begin{table}[t]
\centering\small
\begin{tabular}{@{}lp{0.88\columnwidth}@{}}
\toprule
Pair & Requirement \\
\midrule
A & R1. Every statement of the income used in the affordability or servicing calculation must give the monthly income stated in Employment. \\
A & R2. Every statement of the approved facility limit must give the same amount, and any stated change to the limit must arrive at that amount. \\
B & R3. Every sum the memo writes out about monthly income must be arithmetically correct, whether it adds an amount, deducts one, or compares the income with an earlier figure. \\
B & R4. Every income multiple quoted for the facility limit must follow from the approved limit and the monthly income stated in Employment. \\
C & R5. Every statement about the repayment record must agree with the six periods shown in the repayment history. \\
C & R6. Every stated period of employment or sector experience must be possible given the applicant's age and the age at which education was completed. \\
\bottomrule
\end{tabular}
\caption{The six requirements and the three pairs in which the team receives them.}
\label{tab:1}
\end{table}

Each requirement is labelled on each memo. We call one requirement on one memo a cell, so there are 1,800 cells. A cell is clean when every statement the requirement governs is correct. Otherwise it has one of three kinds of defect. We write $y_{ij} = 1$ when memo $i$ has a defect under requirement $j$ and $y_{ij} = 0$ when the cell is clean. A bare defect is a statement that conflicts with another statement of the same fact, with no reason given. A masked defect comes with its own explanation and is consistent in itself, yet contradicts other facts in the memo. An arithmetic defect is a written-out sum or ratio that is wrong. Under R1, for example, one memo with a bare defect gives monthly income as NT\$23,200 and then computes a ratio ``against monthly income of NT\$46,900''. One with a masked defect gives monthly income as NT\$37,000 and then states that the servicing test ``was run on monthly income of NT\$31,700, being the stated figure net of NT\$5,300 in non-recurring overtime''. There are 1,136 clean cells, and 205 bare, 229 masked and 230 arithmetic defects. That puts a defect in 37 percent of cells, far more than in practice. The median memo has two defects, and 21 memos have none.

A label was accepted when three sources agreed. The first was the generator that set it. The second was a separate program that recomputed it from the finished text. The third was Claude Opus 5.5, a larger model than any we test, which read the memo without seeing either label. All 1,800 labels passed after one revision to the wording of two requirements, which Appendix A describes. Every defect is an inconsistency inside the document that a program can verify. Omissions, errors against outside facts and matters of judgement are not in the test.

\subsection{Arms}

By ``agent'' we mean one call of a model with its own instructions and its own context, and by ``team'' several agents whose verdicts are taken together. Each agent is a single call, so the design can equally be described as a decomposition of a judge prompt, and \citet{schluntz2024building} would class it as a parallel workflow. Our team is the simplest kind, with no orchestrator or filter, and in that respect resembles the architecture that \citet{kim2025scaling} call independent. Four arms review every memo.

Let $R$ be the set of six requirements and $d_i$ memo $i$. A call $f(d_i, S, C)$ gives the model the memo, a set $S \subseteq R$ of requirements to judge and a set $C \subseteq R \setminus S$ of requirements shown for context only. It returns a verdict $v_{ij} \in \{0, 1\}$ for each $j \in S$, where 1 is a flag. The model, the memo, the wording of each requirement and the rest of the prompt are the same in every call. The arms differ in $S$, in $C$ and in the number of calls. They differ otherwise only in the order of the requirement list, given at the end of this section, and in the seeds and, on the Qwen3.5 models, the sampling temperature of the three draws, which Section 3.4 describes.

\[
\begin{array}{@{}ll@{}}
\textrm{Single agent} & f(d_i, R, \emptyset) \\[2pt]
\textrm{Team} & f(d_i, P_m, \emptyset), \quad m = 1, 2, 3 \\[2pt]
\textrm{Scoped team} & f(d_i, P_m, R \setminus P_m), \quad m = 1, 2, 3 \\[2pt]
\textrm{Three draws} & v_{ij} = 1 \textrm{ if } \sum_{s=1}^{3} v_{ij}^{(s)} \ge t
\end{array}
\]

Here $\{P_1, P_2, P_3\}$ is the partition of $R$ into the three pairs of Table~\ref{tab:1}, and a requirement's verdict comes from the call whose $S$ contains it. The pairs are thematic and fixed, and we test this one partition only. The single agent is the partition with one part, and the guide's rule of one consideration to a call is the partition with six.

The team changes two things at once. A team agent holds fewer requirements, and it does not see the four it was not given. The scoped team separates the two by setting $C$ to those four, which appear under a heading that says other reviewers judge them.

For the three draws, $v_{ij}^{(s)}$ is the verdict of the $s$-th sample of the single agent's call and $t$ is the number of flags required. The majority rule is $t = 2$, the any-of-three rule is $t = 1$, and the all-of-three rule, $t = 3$, is used only for the line in Figure~\ref{fig:1}. In the team a single call decides each requirement, so the majority rule is stricter for flagging than the team's.

The order of the requirements in the single agent's list is rotated across memos and across draws. The order within a team call is fixed.

\subsection{Models and instruction}

We use four models from two families. Qwen3.5 at 9B and 27B parameters run locally, and Claude Haiku 5.5 and Claude Sonnet 5.5 through a batch interface. Every comparison is within one model. Answers are returned under a JSON schema that holds, for each requirement, a reasoning string, a verdict of PASS or FAIL and a quoted evidence sentence. Separate thinking channels are switched off, and the output cap is 2,048 tokens for each requirement in every arm. The Qwen3.5 models are sampled at temperature 0, and at 0.7 for the three draws, which would otherwise be near copies of one another. The Claude models accept no temperature setting and use default sampling in every arm. Comparisons between the three draws and the other arms are therefore confounded with temperature on the Qwen3.5 models and not on the Claude models.

All arms carry the same instruction on how to reason. The reviewer is told to take each requirement once, to say in at most three sentences what it compared, and not to return to a requirement after deciding it. We call this the decide-once instruction. We made it the main condition before the runs reported here, for a measurement reason. Left to reason freely in a preliminary run, which Appendix C describes, Qwen3.5 27B returned no verdict on about 6 percent of calls. A failed single-agent call forfeits six verdicts and a failed team call two, so the losses would have fallen unevenly on the arms. The instruction departs from default practice, and it has effects of its own. On 100 of the 300 memos, selected with a fixed seed, every arm is also run without it, and Section 4.4 reports what changes for the single agent and the team.

\subsection{Scoring and analysis}

A flag is a verdict of FAIL, and a false alarm is a flag on a clean cell. For an arm $a$ with verdicts $v_{ij}^{a}$, where $a$ is single, team, scoped, or a rule on the three draws, we report the share of defects caught, $H_a$, and the false-alarm rate on clean cells, $F_a$.

\[
H_a = \frac{\sum_{i,j} y_{ij} \, v_{ij}^{a}}{\sum_{i,j} y_{ij}}, \qquad
F_a = \frac{\sum_{i,j} (1 - y_{ij}) \, v_{ij}^{a}}{\sum_{i,j} (1 - y_{ij})}
\]

Within one class of defect, with both sums restricted to that class, the share caught is written $H_a^{\mathrm{bare}}$, $H_a^{\mathrm{masked}}$ or $H_a^{\mathrm{arith}}$. Balanced accuracy is $B_a = (H_a + 1 - F_a)/2$, which weights the two errors equally whatever their frequency. We write $\Delta H = H_{\mathrm{team}} - H_{\mathrm{single}}$ for the team's value minus the single agent's, and $\Delta F$ and $\Delta B$ likewise. Other contrasts are written out, as in $F_{\mathrm{scoped}} - F_{\mathrm{team}}$. An arm returns verdicts and not scores, so each arm gives a single operating point $(F_a, H_a)$. The sums run over all memos, so rates are pooled over memos and, for all defects, over the three classes. The tokens $K_a$ of an arm are its input and output tokens together.

A missing verdict counts against the arm, and intervals are bootstrap intervals over memos, at 95 percent unless stated. We call a contrast significant when its interval excludes zero. Appendix A gives both rules in full. Each arm was run once, so the intervals cover the sampling of memos and the noise of calls within a run, and do not cover variation between whole runs. Repeated calls exist for the single agent's prompt only, and Table~\ref{tab:A8} gives their spread.

The memos, prompts, predictions and analysis code were fixed by a committed hash before each family was run. Five predictions were set from a pilot, an earlier run of the same comparison on a separate memo set that contributes no number reported here. In the pilot the team's gain appeared on defects that carry an explanation and not on bare ones, which is why the masked contrast is primary. Two were primary, that $\Delta H^{\mathrm{masked}} > 0$ and that $\Delta F > 0$. Each model family has two models, so the two primary predictions give four tests per family. Every primary test is reported at the Bonferroni level of 98.75 percent, which we call the corrected level, and the secondary predictions are tested at 95 percent. At that level the interval on the masked contrast has a half-width of about 0.11 on Qwen3.5 9B, 0.08 on Qwen3.5 27B and 0.07 on Haiku 5.5, so an effect near 0.08 could be detected on some models and not on others. Three predictions were secondary. They were that the team and the single agent do not differ on bare defects, that the team catches more masked defects than a majority of three draws, and that its excess of false alarms is smaller on the larger model of a family.

The scoped team was added after the other arms had been read. Its two predictions, fixed before it was run, were that it raises fewer false alarms than the team and that it still catches more masked defects than the single agent. Neither is a primary prediction of the study. The idea came from reading the Sonnet 5.5 team's false alarms, and the arm is compared with that same team run. We report its contrasts at the corrected level, and Appendix B notes which also hold at 99.375 percent, the level for eight tests per family.

Everything else is exploratory. That covers balanced accuracy, the signal detection analysis of Appendix D, the any-of-three rule and the comparison of arms with the line through the three-draw rules, the effect of the instruction, the break-even costs, and the rates by requirement and by memo. Appendix B lists every prediction with its outcome.

\section{Results}

\begin{figure*}[t]
\centering
\includegraphics[width=0.82\textwidth]{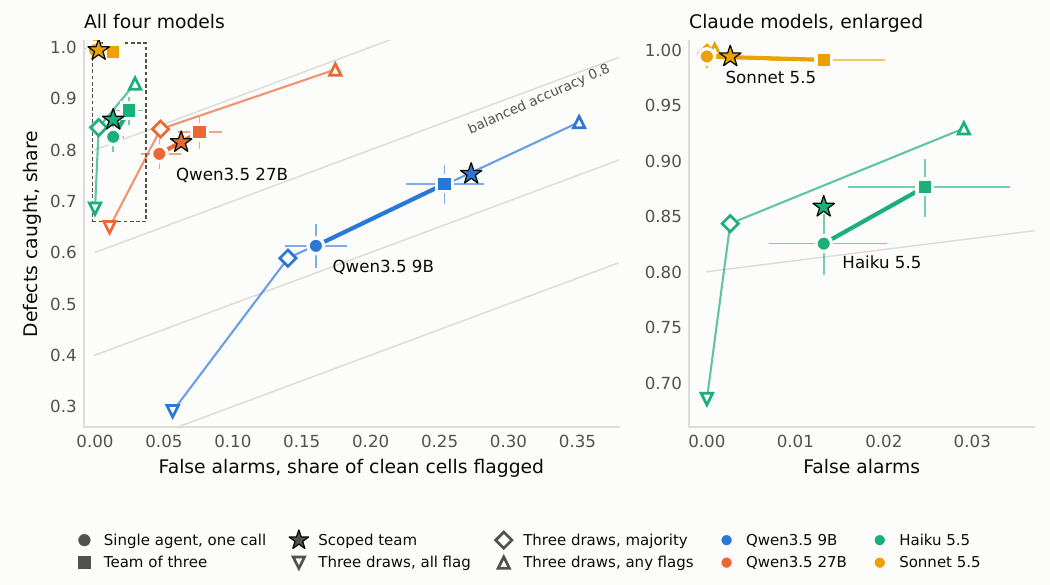}
\caption{Each arm as an operating point, false alarms on clean cells against the share of all defects caught, with the Claude models enlarged on the right. A thick segment joins a model's single agent to its team, and a star marks the scoped team. A thin line joins three rules applied to three draws of the single agent's prompt, flagging when all, at least two or any of the draws flag. Bars on the single agent and the team are 95 percent intervals. Grey lines join points of equal balanced accuracy.}
\label{fig:1}
\end{figure*}

No arm on any model lost more than two calls under the decide-once instruction. Table~\ref{tab:2} reports the rates and Figure~\ref{fig:1} shows each arm as an operating point. The team uses 2.2 to 2.4 times the single agent's tokens, and the three draws use more.

\begin{table*}[t]
\centering\small
\begin{tabular}{@{}l@{\hspace{5pt}}l@{\hspace{5pt}}r@{\hspace{6pt}}r@{\hspace{6pt}}r@{\hspace{6pt}}r@{\hspace{6pt}}r@{\hspace{6pt}}r@{\hspace{6pt}}>{\raggedright\arraybackslash}p{0.1\textwidth}@{}}
\toprule
Model & Arm & $H^{\mathrm{bare}}$ & $H^{\mathrm{masked}}$ & $H^{\mathrm{arith}}$ & $H$ & $F$ & $B$ & $K_a / K_{\mathrm{single}}$ \\
\midrule
Qwen3.5 9B & Single agent & 0.717 & 0.559 & 0.574 & 0.613 & 0.160 & 0.726 & 1.00 \\
 & Team & 0.849 & 0.642 & 0.722 & 0.733 & 0.254 & 0.740 & 2.36 \\
 & Scoped team & 0.898 & 0.646 & 0.730 & 0.753 & 0.273 & 0.740 & 2.55 \\
 & Three draws, majority & 0.756 & 0.402 & 0.626 & 0.589 & 0.140 & 0.724 & 3.11 \\
 & Three draws, any & 0.902 & 0.795 & 0.870 & 0.854 & 0.351 & 0.751 & 3.11 \\
Qwen3.5 27B & Single agent & 0.971 & 0.734 & 0.691 & 0.792 & 0.047 & 0.873 & 1.00 \\
 & Team & 1.000 & 0.817 & 0.704 & 0.834 & 0.076 & 0.879 & 2.24 \\
 & Scoped team & 0.995 & 0.808 & 0.661 & 0.815 & 0.062 & 0.876 & 2.40 \\
 & Three draws, majority & 1.000 & 0.803 & 0.735 & 0.840 & 0.048 & 0.896 & 3.06 \\
 & Three draws, any & 1.000 & 0.961 & 0.913 & 0.956 & 0.174 & 0.891 & 3.06 \\
Haiku 5.5 & Single agent & 0.946 & 0.790 & 0.752 & 0.825 & 0.013 & 0.906 & 1.00 \\
 & Team & 1.000 & 0.891 & 0.752 & 0.877 & 0.025 & 0.926 & 2.16 \\
 & Scoped team & 0.990 & 0.882 & 0.717 & 0.858 & 0.013 & 0.923 & 2.30 \\
 & Three draws, majority & 0.966 & 0.803 & 0.774 & 0.843 & 0.003 & 0.920 & 3.00 \\
 & Three draws, any & 0.995 & 0.908 & 0.891 & 0.929 & 0.029 & 0.950 & 3.00 \\
Sonnet 5.5 & Single agent & 1.000 & 1.000 & 0.983 & 0.994 & 0.000 & 0.997 & 1.00 \\
 & Team & 1.000 & 0.996 & 0.978 & 0.991 & 0.013 & 0.989 & 2.17 \\
 & Scoped team & 1.000 & 1.000 & 0.983 & 0.994 & 0.003 & 0.996 & 2.32 \\
 & Three draws, majority & 1.000 & 1.000 & 0.991 & 0.997 & 0.000 & 0.998 & 3.00 \\
 & Three draws, any & 1.000 & 1.000 & 1.000 & 1.000 & 0.001 & 1.000 & 3.00 \\
\bottomrule
\end{tabular}
\caption{Share of defects caught by class and over all defects, $H$, false alarms on clean cells, $F$, balanced accuracy, $B$, and tokens relative to the single agent. The majority rule flags when at least two of three draws flag, and the any rule when at least one does. All 300 memos, decide-once instruction.}
\label{tab:2}
\end{table*}

\subsection{The team flags more, and catches more below ceiling}

\begin{figure*}[t]
\centering
\includegraphics[width=0.92\textwidth]{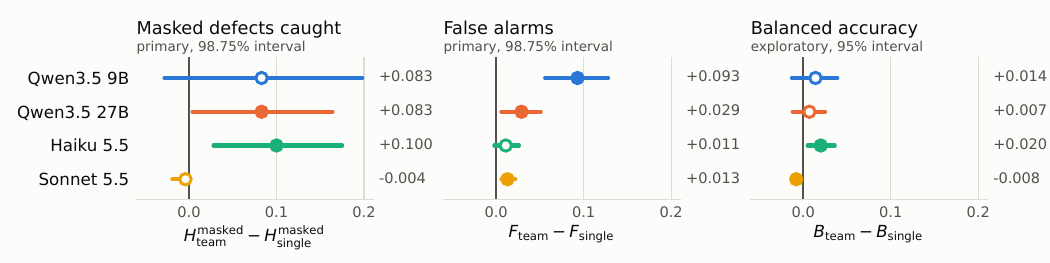}
\caption{The team against the single agent, on one scale. The contrasts in masked defects caught, $\Delta H^{\mathrm{masked}}$, and in false alarms, $\Delta F$, are primary, with intervals at the corrected level. The contrast in balanced accuracy, $\Delta B$, is exploratory, with 95 percent intervals. A filled marker has an interval that excludes zero, and the number beside each row is the estimate. Table~\ref{tab:A1} gives the bounds.}
\label{fig:2}
\end{figure*}

On the three models whose single agent leaves masked defects uncaught, the team's share caught is higher by 0.08 to 0.10, as Figure~\ref{fig:2} shows. The prediction is confirmed at the corrected level on Qwen3.5 27B and Haiku 5.5 and not on Qwen3.5 9B, where the estimate equals that on Qwen3.5 27B and the interval is wider. The team raises more false alarms on every model, at the corrected level on three. The secondary prediction of no difference on bare defects was wrong. On the three models below ceiling the team catches more of them, as Appendix B records. In Figure~\ref{fig:1} the segment from single agent to team rises to the right on those models.

With the two errors weighted equally the effects largely cancel. Balanced accuracy changes by 0.02 or less on every model, and all four intervals lie within 0.04 of zero. We set no equivalence margin in advance, and the change is significant on the two Claude models, a rise on Haiku 5.5 and a fall on Sonnet 5.5. Sonnet 5.5's single agent catches every bare and masked defect, and its team catches two fewer defects and adds 15 false alarms.

As predicted for the larger model of a family, the team's excess of false alarms is smaller on Qwen3.5 27B than on Qwen3.5 9B, as Appendix B records. In proportion to each model's own false-alarm rate the two excesses are about the same, and the two Claude models do not differ.

The pooled rates conceal changes in opposite directions. The team's share of defects caught rises on some requirements and falls on others, as Table~\ref{tab:A7} shows, so the trade describes the review as a whole and not each part of it. On the pooled rates, an exploratory signal detection analysis in Appendix D is consistent with a lower threshold for flagging and an unchanged ability to tell defective cells from clean ones, and does not establish that account.

\subsection{Scope information}

We read the 15 false alarms of the Sonnet 5.5 team. Most fall on requirement R1. The agent holding R1 flags the sentence that adds a housing allowance to monthly income. That sentence falls under R3, a requirement this agent was not shown. The single agent holds both and does not make the error. By the rates in Table~\ref{tab:A7}, R1 and R3 account for about 0.8 of the team's added false alarms on Sonnet 5.5 and Qwen3.5 27B. On Haiku 5.5 most of the addition falls on R4, and on Qwen3.5 9B it is spread across five requirements. The scoped team restores to each agent the requirements it was not shown.

\begin{figure*}[t]
\centering
\includegraphics[width=0.92\textwidth]{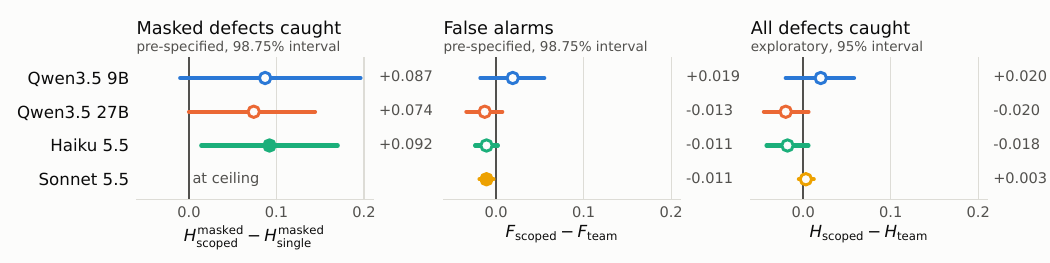}
\caption{The scoped team, on the scale of Figure~\ref{fig:2}. Its contrast with the single agent in masked defects caught and its contrast with the team in false alarms are the two pre-specified contrasts, with intervals at the corrected level. Its contrast with the team in all defects caught is exploratory, with 95 percent intervals. Markers and numbers are as in Figure~\ref{fig:2}. Table~\ref{tab:A2} gives the bounds.}
\label{fig:3}
\end{figure*}

The scoped team raises fewer false alarms than the team on Sonnet 5.5, at the corrected level, as Figure~\ref{fig:3} shows. The reduction on Haiku 5.5 is the same size and is significant at 95 percent, [$-$0.021, $-$0.002]. On Qwen3.5 27B the reduction is not significant, and on Qwen3.5 9B there is none. The scoped team still catches more masked defects than the single agent on Haiku 5.5. That prediction is not confirmed on Qwen3.5 9B, and on Qwen3.5 27B the interval reaches zero.

The reduction may not be free. On Qwen3.5 27B and Haiku 5.5 the scoped team's share of all defects caught is about 0.02 below the team's, mostly through arithmetic defects. On those two models each 0.01 taken off the false-alarm rate comes with a fall of about 0.015 in the share of defects caught, which is not significant. Sonnet 5.5 gives up nothing.

Scope information may act by telling the agent what lies outside its requirements, or by making it more cautious because other reviewers are named. Our prompt does not separate the two. In Table~\ref{tab:2}, Haiku 5.5's scoped team has the single agent's false-alarm rate, catches more masked defects and catches fewer arithmetic defects. The scoped prompt adds about 160 to 250 input tokens a call to the team's.

\subsection{Repeated draws}

The remaining secondary prediction was that the team catches more masked defects than a majority of three draws. It is confirmed on Haiku 5.5 and not on Qwen3.5 27B or Sonnet 5.5. On Qwen3.5 9B the estimate is above zero, but on the Qwen3.5 models the comparison is confounded with temperature. Appendix B gives the estimates. A draw of the 9B model catches fewer masked defects than its single agent at temperature 0, as Table~\ref{tab:A8} shows.

Under the exploratory any-of-three rule the three-draw agent catches more defects than the team on all four models, and its balanced accuracy is no lower than the team's on any. On the two Claude models it is higher by 0.024 and 0.011, both significant.

A further exploratory check, unlike the signal detection analysis of Appendix D, makes no distributional assumption. It compares each arm with the line in Figure~\ref{fig:1} that joins the three rules on three draws, taken at the arm's own false-alarm rate. Haiku 5.5's team lies 0.038 below the line [$-$0.065, 0.003], and its single agent 0.052 below [$-$0.093, $-$0.018]. On that model repeated draws therefore attain operating points that one call does not, and probably points that the team does not. On the Qwen3.5 models the team is 0.002 above the line on the 9B model and 0.032 below it on the 27B model, neither significant. The check cannot be made on Sonnet 5.5, which is at ceiling.

The draws also show how much repeated calls of one prompt vary. They disagree on 0.39 of cells on Qwen3.5 9B, 0.21 on Qwen3.5 27B, 0.11 on Haiku 5.5 and under 0.01 on Sonnet 5.5. By the repeats in Table~\ref{tab:A8}, the team's share of masked defects caught on Haiku 5.5 lies outside the range of four calls of the single agent's prompt, while on Qwen3.5 27B three draws vary as much as the team's gain. The single agent's share of defects caught also differs across the three rotations of its list, by up to 0.14 on Qwen3.5 9B, 0.09 on Qwen3.5 27B and 0.05 on Haiku 5.5. Each rotation covers a different 100 memos, so the order of the list and the memos are not separated.

\subsection{The decide-once instruction, exploratory}

On the 100 memos run with and without the instruction, removing it raises the single agent's share of defects caught. With calls that share a prompt and settings combined, masked defects caught rise significantly on the three models below ceiling, as Table~\ref{tab:A3} shows. Haiku 5.5's false alarms do not rise with them. For the Qwen3.5 models these combined calls are the draws at temperature 0.7. In the single call at temperature 0, which Table~\ref{tab:A3} also reports, arithmetic defects caught rise on both models and masked defects caught do not rise significantly on either.

The effect of division also depends on the instruction, as Table~\ref{tab:3} shows. Without it the team's excess of false alarms is larger on all four models, significantly on Qwen3.5 27B, +0.088 [0.013, 0.161], and Sonnet 5.5, +0.013 [0.003, 0.026]. The team's balanced accuracy is then significantly lower than the single agent's on Qwen3.5 27B and Sonnet 5.5 and significantly higher on Haiku 5.5. On Qwen3.5 9B the difference is not significant. On Qwen3.5 27B the team's gain in defects caught disappears while its excess of false alarms grows. Team calls cut off at the output cap account for part of this, as Appendix C records. On completed verdicts alone the team still raises more false alarms and catches almost no more defects.

\begin{table*}[t]
\centering\small\setlength{\tabcolsep}{4pt}
\begin{tabular}{@{}l>{\raggedright\arraybackslash}p{0.13\textwidth}>{\raggedright\arraybackslash}p{0.13\textwidth}>{\raggedright\arraybackslash}p{0.13\textwidth}>{\raggedright\arraybackslash}p{0.13\textwidth}>{\raggedright\arraybackslash}p{0.13\textwidth}>{\raggedright\arraybackslash}p{0.13\textwidth}@{}}
\toprule
Model & $\Delta H$, with & $\Delta H$, without & $\Delta F$, with & $\Delta F$, without & $\Delta B$, with & $\Delta B$, without \\
\midrule
Qwen3.5 9B & +0.133\newline \mbox{\footnotesize [0.050, 0.214]} & +0.084\newline \mbox{\footnotesize [0.000, 0.164]} & +0.110\newline \mbox{\footnotesize [0.065, 0.157]} & +0.126\newline \mbox{\footnotesize [0.072, 0.179]} & +0.012\newline \mbox{\footnotesize [$-$0.037, 0.060]} & $-$0.021\newline \mbox{\footnotesize [$-$0.068, 0.026]} \\
Qwen3.5 27B & +0.071\newline \mbox{\footnotesize [0.017, 0.126]} & +0.004\newline \mbox{\footnotesize [$-$0.046, 0.052]} & +0.035\newline \mbox{\footnotesize [0.006, 0.064]} & +0.123\newline \mbox{\footnotesize [0.050, 0.192]} & +0.018\newline \mbox{\footnotesize [$-$0.013, 0.049]} & $-$0.059\newline \mbox{\footnotesize [$-$0.105, $-$0.013]} \\
Haiku 5.5 & +0.022\newline \mbox{\footnotesize [$-$0.022, 0.066]} & +0.071\newline \mbox{\footnotesize [0.026, 0.118]} & +0.008\newline \mbox{\footnotesize [$-$0.008, 0.026]} & +0.019\newline \mbox{\footnotesize [0.000, 0.038]} & +0.007\newline \mbox{\footnotesize [$-$0.016, 0.030]} & +0.026\newline \mbox{\footnotesize [0.001, 0.053]} \\
Sonnet 5.5 & $-$0.004\newline \mbox{\footnotesize [$-$0.019, 0.009]} & $-$0.013\newline \mbox{\footnotesize [$-$0.030, 0.000]} & +0.008\newline \mbox{\footnotesize [0.000, 0.018]} & +0.021\newline \mbox{\footnotesize [0.008, 0.037]} & $-$0.006\newline \mbox{\footnotesize [$-$0.015, 0.002]} & $-$0.017\newline \mbox{\footnotesize [$-$0.028, $-$0.008]} \\
\bottomrule
\end{tabular}
\caption{The contrasts $\Delta H$ over all defects, $\Delta F$ and $\Delta B$ on the 100 paired memos, with the decide-once instruction and without it. 95 percent intervals.}
\label{tab:3}
\end{table*}

Part of the team's gain on the Qwen3.5 models under the instruction may be a recovery of detection that the instruction takes from the single agent. A direct comparison on these 100 memos, the single agent without the instruction against the team with it, does not settle the question, as Table~\ref{tab:A4} shows.

\subsection{What the trade costs, exploratory}

Table~\ref{tab:4} expresses the trade of Section 4.1 as the rise in the false-alarm rate for each unit rise in the share of all defects caught. Whether the exchange is worthwhile depends on how common defects are. Let $p$ be the share of cells that have a defect, $w_M$ the cost of a missed defect and $w_F$ the cost of a false alarm. With cost taken as linear in the two errors, arm $a$ has an expected cost of errors per cell of

\[
E_a = p \, (1 - H_a) \, w_M + (1 - p) \, F_a \, w_F .
\]

With $\Delta H > 0$ over all defects, the team has the lower expected cost when

\[
\frac{w_M}{w_F} > \frac{\Delta F}{\Delta H} \cdot \frac{1 - p}{p} .
\]

\begin{table*}[t]
\centering\small
\begin{tabular}{@{}ll@{\hspace{14pt}}rrrr@{}}
\toprule
Model & $\Delta F / \Delta H$ & $p = 0.37$ & $p = 0.05$ & $p = 0.02$ & $p = 0.01$ \\
\midrule
Qwen3.5 9B & 0.77 [0.48, 1.33] & 1.3 & 15 & 38 & 77 \\
Qwen3.5 27B & 0.69 [0.21, 2.73] & 1.2 & 13 & 34 & 68 \\
Haiku 5.5 & 0.22 [0.01, 0.65] & 0.4 & 4 & 11 & 22 \\
\bottomrule
\end{tabular}
\caption{Team against single agent under the decide-once instruction. The second column is $\Delta F / \Delta H$, with $\Delta H$ over all defects and a 95 percent interval. The others are the break-even value of $w_M / w_F$ at four values of the defect share $p$, assuming the rates carry over.}
\label{tab:4}
\end{table*}

At the defect share of our memos the break-even cost is between 0.4 and 1.3, and at shares of 0.05 to 0.01 it is roughly 10 to 60 times higher. The ratios are imprecise. At a share of 0.02 the interval on Qwen3.5 27B alone gives break-even costs from 10 to 134. Sonnet 5.5 has no row, since its team catches no more than its single agent.

Two cautions apply to Table~\ref{tab:4}. It uses the false-alarm rate by cell, which probably overstates the team's cost, as Section 5 explains. It also compares the team with the single agent only. We set the five rows that Table~\ref{tab:2} gives for each model side by side at a defect share of 0.02, for costs of a miss from 1 to 200 times that of a false alarm. A majority of three draws has the lowest $E_a$ while a miss costs less than roughly 15 to 55 false alarms, depending on the model. Above that the any-of-three rule has the lowest. The team has the lowest expected cost at none of the ratios we tried, on any model. On Qwen3.5 9B it lies on the line joining the two three-draw rules, so the ranking there is not robust. Table~\ref{tab:A10} reports the share of flags that would be correct at three defect shares.

The unit of work in practice is the file. The share of the 300 memos with at least one false alarm is higher for the team than for the single agent on every model, by 0.04 to 0.19. Table~\ref{tab:A6} reports related counts for whole memos.

\section{Threats to validity and follow-up studies}

Four threats could change a conclusion. For the first three we give the study that would address it, and we rank those studies by how far they could change these conclusions.

\textbf{How well the effects are measured.} On Haiku 5.5 most of the team's flags on clean cells quote a sentence that is defective under another requirement, and one of the single agent's does. We call these misfiled flags, and Table~\ref{tab:A9} counts them for every model. The count rests on an exploratory match of text, supported by a hand check of twelve flags. The team's cost in false alarms is therefore probably overstated on Qwen3.5 27B, Haiku 5.5 and Sonnet 5.5. Its gain on masked defects is less secure on Qwen3.5 27B, where three draws of one prompt vary as much as the gain, than on Haiku 5.5, where the team lies outside the range of four calls. A study that scores flags by what they quote, repeats its arms and samples the three draws at the temperature of the other arms would establish whether the false-alarm cost that defines the trade is real. We rank it second.

\textbf{The instruction.} Every primary test was run under the decide-once instruction, and Section 4.4 shows that on Qwen3.5 27B the team's balanced accuracy does not differ significantly from the single agent's with the instruction and is lower without it. The primary results hold for one way of telling the reviewer to reason. Crossing the division with the instruction on the full memo set is the study we rank first, since this is the only threat for which our own data show a result changing.

\textbf{The division.} We test one pairing, two requirements to a call. R1 and R3 fall in different pairs, and Section 4.2 traces most of the team's added false alarms on two models to them. The effect sizes are specific to this pairing, and the guide's rule as stated, one consideration to a call, is untested. Varying the division, down to one requirement a call, would test it and ranks third.

\textbf{Generality.} The memos are short and synthetic, with internal inconsistencies in 37 percent of cells, and the strongest model is at ceiling. Neither the rates nor the costs should be assumed to carry over to long documents with rare and varied defects, where outcomes should also be reported by file.

\section{Guidance for agent setup}

Whether a division of this kind helps depends on what the single agent misses, on the model, and on how the reviewer is told to reason. Three implications for the design of a review system follow, ordered by the strength of the evidence behind them. All three assume that rates measured on these memos carry over to the designer's documents, which we have not shown.

The first rests on the primary comparison and holds under the decide-once instruction for the pairing we tested. Division should be treated as a change in what is flagged and not as an improvement, and its cost in false alarms should be worked out before it is adopted. The gain is bounded by what the single agent misses, and the false alarms are not, as the Sonnet 5.5 rows of Table~\ref{tab:2} show. A designer therefore needs the single agent's miss rate and false-alarm rate on cases known to be clean, and the defect share of the setting. With those, the exploratory calculation behind Table~\ref{tab:4} gives the cost of a miss above which division is worthwhile relative to a single call.

The second rests on an exploratory comparison that is free of the temperature confound on the two Claude models only. The single agent sampled three times is the baseline against which a divided review should be judged, since it uses the same number of calls. On those models the any-of-three rule gives higher balanced accuracy than the team, and a majority of three raises no more false alarms than one call, at a higher token cost than the team, as Table~\ref{tab:2} shows. Which rule suits a setting depends, as before, on the cost of a miss.

The third rests on 100 memos and is the least secure. How the reviewer is told to reason should be fixed and reported before arrangements are compared, because the instruction changed the effect of division. On Qwen3.5 27B the team's balanced accuracy does not differ significantly from the single agent's with the instruction and is lower without it, as Table~\ref{tab:3} shows. Removing a constraint on reasoning also risks calls that return no verdict.

We recommend one practice whatever the follow-up studies of Section 5 find. A divided review should be judged on misses and false alarms separately, on cases known to be clean, and against the single agent given the same number of calls.

\section{Conclusion}

Under the decide-once instruction, dividing six requirements among three agents on one model trades false alarms for defects caught, and with the two errors weighted equally the effects largely cancel, as Figure~\ref{fig:2} shows. The trade of more defects caught for more false alarms is what \citet{he2026fined} found with the whole document in the prompt and what \citet{liu2026dude} found for a team with no filter, and the small change in accuracy is what \citet{rao2026jev} found for binary checklists. In the terms of \citet{fox2026omission}, the exploratory signal detection analysis of Appendix D is consistent with division sliding the operating point and does not establish it. Repeated draws behave differently in our data from theirs. On Haiku 5.5 three draws attain operating points that one call does not, as Section 4.3 shows. Their task and their voting rule differ from ours, and we have not tested which of the two explains the disagreement.

Besides the exploratory comparisons, two things are new. The false-alarm rate is measured on cells known to be clean, which lets the trade be sized. The scoped team, an arm added after the first results, raises fewer false alarms than the team on the two Claude models, at the corrected level on one.

\bibliography{references}
\appendix
\renewcommand{\topfraction}{0.95}\renewcommand{\bottomfraction}{0.95}\renewcommand{\textfraction}{0.03}\renewcommand{\floatpagefraction}{0.9}\setcounter{topnumber}{4}\setcounter{bottomnumber}{4}\setcounter{totalnumber}{8}
\setcounter{table}{0}
\renewcommand{\thetable}{A\arabic{table}}

\section{Memos, labels and settings}

\textbf{Memos.} The record supplies the credit limit, age, education, marital status and six months of repayment status, bills and payments from April to September 2005. The dataset has no income or employment fields. The program draws monthly income, allowances, tenure and sector experience for each record from fixed ranges tied to the credit limit and age, and writes them into sentences from templates. Every figure a requirement tests is stated only by sentences the program controls. A defect changes one sentence, and no other requirement governs that sentence. An agent may still judge the sentence against a requirement of its own. The label of each requirement is drawn independently on each memo.

\textbf{Labels.} In a first pass the third source, Claude Opus 5.5, differed from the other two on five cells. In four it found a figure wrong and doubted whether the sentence fell under the requirement or what the figure referred to, and in the fifth it found no error. The wording of two requirements was made explicit, the memos were regenerated with one sentence changed, and the two affected requirements were read again on all 300 memos, with agreement on every cell. No memo was dropped. Opus 5.5 belongs to the family of two of the tested models and is not one of them.

\textbf{Models and settings.} The Qwen3.5 models ran under Ollama 0.35.1 with 4-bit quantisation, Q4\_K\_M, a context of 16,384 tokens and a fixed seed, and the three draws used three seeds. The Claude models ran through the Batch API with effort set to high. Runs took place between 8 and 10 October 2026. The scoped team was run one day after the other arms. The first requirement in the single agent's list is rotated by memo index across the three pairs, and each draw advances the rotation by one.

\textbf{Scoring rules.} A missing verdict counts as a miss on a defect and as a flag on a clean cell. A defect counts as caught when the verdict is FAIL, whatever sentence the agent quotes. Intervals are percentile bootstrap intervals from 5,000 resamples of memos, with arms paired by memo.

\textbf{The decide-once instruction.} The sentences below are added to the prompt of every arm.

\begin{quote}\ttfamily\footnotesize\raggedright\setlength{\parindent}{0pt}
Take each requirement once. In at most three sentences say what you compared and what it shows, then give the verdict. Do not go back over a requirement once you have decided it.\par
\end{quote}

\textbf{The scoped prompt.} A scoped call lists the requirements under two headings and ends the second list with a note. Text in angle brackets stands for the wording of the requirements.

\begin{quote}\ttfamily\footnotesize\raggedright\setlength{\parindent}{0pt}
REQUIREMENTS YOU JUDGE\par
\textless{}the two requirements of this call\textgreater{}\par
\vspace{4pt}
REQUIREMENTS JUDGED BY OTHER REVIEWERS\par
\textless{}the other four requirements\textgreater{}\par
Do not give a verdict on the requirements in this second list. Other reviewers judge them. They are listed so that you know what they cover.\par
\end{quote}

\section{Predictions and outcomes}

The seven predictions follow, in three groups, each with its outcome.

\noindent \textbf{Primary, tested at the 98.75 percent level.}

{\footnotesize\setlength{\parindent}{0pt}\hangindent=1em\hangafter=1 \textit{Prediction.} The team catches more masked defects than the single agent.\par\hangindent=1em\hangafter=1 \textit{Outcome.} Holds on Qwen3.5 27B and Haiku 5.5. Not shown on Qwen3.5 9B. No room on Sonnet 5.5.\par}\vspace{3pt}

{\footnotesize\setlength{\parindent}{0pt}\hangindent=1em\hangafter=1 \textit{Prediction.} The team raises more false alarms than the single agent.\par\hangindent=1em\hangafter=1 \textit{Outcome.} Holds on Qwen3.5 9B, Qwen3.5 27B and Sonnet 5.5. Not shown at the corrected level on Haiku 5.5.\par}\vspace{3pt}

\noindent \textbf{Secondary, tested at the 95 percent level.}

{\footnotesize\setlength{\parindent}{0pt}\hangindent=1em\hangafter=1 \textit{Prediction.} The team and the single agent do not differ on bare defects.\par\hangindent=1em\hangafter=1 \textit{Outcome.} Wrong. The team catches more on three models, by 0.132 [0.062, 0.202], 0.029 [0.009, 0.055] and 0.054 [0.025, 0.086].\par}\vspace{3pt}

{\footnotesize\setlength{\parindent}{0pt}\hangindent=1em\hangafter=1 \textit{Prediction.} The team catches more masked defects than a majority of three draws.\par\hangindent=1em\hangafter=1 \textit{Outcome.} Holds on Haiku 5.5, 0.087 [0.041, 0.133]. Confounded with temperature on Qwen3.5 9B, 0.240 [0.153, 0.328]. Not shown on Qwen3.5 27B, 0.013 [$-$0.048, 0.076], or Sonnet 5.5.\par}\vspace{3pt}

{\footnotesize\setlength{\parindent}{0pt}\hangindent=1em\hangafter=1 \textit{Prediction.} The team's excess of false alarms is smaller on the larger model of a family.\par\hangindent=1em\hangafter=1 \textit{Outcome.} Holds for Qwen3.5 as a difference, 0.064 [0.030, 0.099]. As a proportion of each model's rate the excess is about the same. Not shown for Claude, where the difference is $-$0.002 [$-$0.015, 0.011].\par}\vspace{3pt}

\noindent \textbf{Pre-specified, added arm, tested at the 98.75 percent level.} The level for eight tests per family is 99.375 percent.

{\footnotesize\setlength{\parindent}{0pt}\hangindent=1em\hangafter=1 \textit{Prediction.} The scoped team raises fewer false alarms than the team.\par\hangindent=1em\hangafter=1 \textit{Outcome.} Holds on Sonnet 5.5, also at the eight-test level. At 95 percent only on Haiku 5.5. Not shown on the Qwen3.5 models.\par}\vspace{3pt}

{\footnotesize\setlength{\parindent}{0pt}\hangindent=1em\hangafter=1 \textit{Prediction.} The scoped team catches more masked defects than the single agent.\par\hangindent=1em\hangafter=1 \textit{Outcome.} Holds on Haiku 5.5, also at the eight-test level. Lower bound 0.000 on Qwen3.5 27B. Not shown on Qwen3.5 9B. No room on Sonnet 5.5.\par}\vspace{3pt}

\begin{table}[!htb]\centering
\footnotesize\setlength{\tabcolsep}{3pt}\renewcommand{\arraystretch}{0.88}
\begin{tabular}{@{}l>{\raggedright\arraybackslash}p{0.235\columnwidth}>{\raggedright\arraybackslash}p{0.235\columnwidth}>{\raggedright\arraybackslash}p{0.235\columnwidth}@{}}
\toprule
Model & $\Delta H^{\mathrm{masked}}$ & $\Delta F$ & $\Delta B$ \\
\midrule
Qwen3.5 9B & +0.083\newline \mbox{\scriptsize [$-$0.028, 0.198]} & +0.093\newline \mbox{\scriptsize [0.056, 0.128]} & +0.014\newline \mbox{\scriptsize [$-$0.013, 0.039]} \\
Qwen3.5 27B & +0.083\newline \mbox{\scriptsize [0.004, 0.164]} & +0.029\newline \mbox{\scriptsize [0.006, 0.051]} & +0.007\newline \mbox{\scriptsize [$-$0.012, 0.025]} \\
Haiku 5.5 & +0.100\newline \mbox{\scriptsize [0.028, 0.175]} & +0.011\newline \mbox{\scriptsize [$-$0.002, 0.026]} & +0.020\newline \mbox{\scriptsize [0.005, 0.036]} \\
Sonnet 5.5 & $-$0.004\newline \mbox{\scriptsize [$-$0.019, 0.000]} & +0.013\newline \mbox{\scriptsize [0.006, 0.022]} & $-$0.008\newline \mbox{\scriptsize [$-$0.013, $-$0.003]} \\
\bottomrule
\end{tabular}
\caption{The contrasts of Figure~\ref{fig:2}, computed from unrounded rates. $\Delta H^{\mathrm{masked}}$ and $\Delta F$ are the two primary contrasts, with intervals at the corrected level. $\Delta B$ is exploratory, with 95 percent intervals.}
\label{tab:A1}
\end{table}

\begin{table}[!htb]\centering
\footnotesize\setlength{\tabcolsep}{3pt}\renewcommand{\arraystretch}{0.88}
\begin{tabular}{@{}l>{\raggedright\arraybackslash}p{0.235\columnwidth}>{\raggedright\arraybackslash}p{0.235\columnwidth}>{\raggedright\arraybackslash}p{0.235\columnwidth}@{}}
\toprule
Model & $H^{\mathrm{masked}}_{\mathrm{scoped}} - H^{\mathrm{masked}}_{\mathrm{single}}$ & $F_{\mathrm{scoped}} - F_{\mathrm{team}}$ & $H_{\mathrm{scoped}} - H_{\mathrm{team}}$ \\
\midrule
Qwen3.5 9B & +0.087\newline \mbox{\scriptsize [$-$0.010, 0.196]} & +0.019\newline \mbox{\scriptsize [$-$0.018, 0.055]} & +0.020\newline \mbox{\scriptsize [$-$0.020, 0.058]} \\
Qwen3.5 27B & +0.074\newline \mbox{\scriptsize [0.000, 0.144]} & $-$0.013\newline \mbox{\scriptsize [$-$0.034, 0.007]} & $-$0.020\newline \mbox{\scriptsize [$-$0.045, 0.006]} \\
Haiku 5.5 & +0.092\newline \mbox{\scriptsize [0.014, 0.170]} & $-$0.011\newline \mbox{\scriptsize [$-$0.024, 0.002]} & $-$0.018\newline \mbox{\scriptsize [$-$0.042, 0.006]} \\
Sonnet 5.5 & at ceiling & $-$0.011\newline \mbox{\scriptsize [$-$0.019, $-$0.003]} & +0.003\newline \mbox{\scriptsize [$-$0.005, 0.012]} \\
\bottomrule
\end{tabular}
\caption{The contrasts of Figure~\ref{fig:3} for the scoped team, computed from unrounded rates. The first two are its pre-specified contrasts, with intervals at the corrected level. The last is exploratory, with 95 percent intervals.}
\label{tab:A2}
\end{table}

\section{The decide-once instruction}

Table~\ref{tab:A3} gives the change in the single agent's rates when the instruction is removed. Without the instruction the Qwen3.5 27B team lost 17 of 300 calls to the output cap, its single agent 2 of 100 and its scoped team 7 of 300. The Qwen3.5 9B team lost 9 of 300. No Claude call was lost. On completed verdicts alone and without the instruction, $\Delta H$ is +0.008 and $\Delta F$ is +0.079 on Qwen3.5 27B, and they are +0.094 and +0.095 on Qwen3.5 9B.

Table~\ref{tab:A4} bears on whether part of the team's gain under the instruction is a recovery of detection that the instruction takes from the single agent. It sets the single agent without the instruction against the team with it. No difference in defects caught is significant, and the single agent without the instruction raises fewer false alarms than the team with it on Qwen3.5 9B and more on Qwen3.5 27B. One hundred memos give wide intervals.

In the preliminary run that led to the instruction, Qwen3.5 27B kept revising its reasoning until it reached the output cap. The alternative to the instruction was to score completed verdicts only, which would have set aside the rule, fixed in advance, that a missing verdict counts against the arm.

\begin{table}[!htb]\centering
\footnotesize\setlength{\tabcolsep}{3pt}\renewcommand{\arraystretch}{0.88}
\begin{tabular}{@{}>{\raggedright\arraybackslash}p{0.26\columnwidth}>{\raggedright\arraybackslash}p{0.215\columnwidth}>{\raggedright\arraybackslash}p{0.215\columnwidth}>{\raggedright\arraybackslash}p{0.215\columnwidth}@{}}
\toprule
Model & $H^{\mathrm{masked}}$ & $H^{\mathrm{arith}}$ & $F$ \\
\midrule
Qwen3.5 9B & +0.179\newline \mbox{\scriptsize [0.105, 0.256]} & +0.163\newline \mbox{\scriptsize [0.093, 0.230]} & +0.046\newline \mbox{\scriptsize [0.012, 0.082]} \\
Qwen3.5 27B & +0.058\newline \mbox{\scriptsize [0.004, 0.112]} & +0.088\newline \mbox{\scriptsize [0.018, 0.159]} & +0.020\newline \mbox{\scriptsize [0.000, 0.039]} \\
Haiku 5.5 & +0.075\newline \mbox{\scriptsize [0.027, 0.123]} & +0.044\newline \mbox{\scriptsize [0.009, 0.079]} & +0.002\newline \mbox{\scriptsize [$-$0.005, 0.009]} \\
Sonnet 5.5 & 0.000 & +0.003\newline \mbox{\scriptsize [$-$0.010, 0.017]} & 0.000 \\
Qwen3.5 9B, single call\newline @Temp=0 & $-$0.037\newline \mbox{\scriptsize [$-$0.162, 0.086]} & +0.188\newline \mbox{\scriptsize [0.065, 0.313]} & +0.005\newline \mbox{\scriptsize [$-$0.039, 0.051]} \\
Qwen3.5 27B, single call\newline @Temp=0 & +0.062\newline \mbox{\scriptsize [$-$0.045, 0.169]} & +0.175\newline \mbox{\scriptsize [0.076, 0.284]} & +0.094\newline \mbox{\scriptsize [0.052, 0.140]} \\
\bottomrule
\end{tabular}
\caption{Change in the single agent's rates when the instruction is removed, the rate without it minus the rate with it, on the 100 paired memos. For the Claude models the single call and the three draws are pooled, and for the Qwen3.5 models the three draws at temperature 0.7. The last two rows give the single call of the Qwen3.5 models at temperature 0. 95 percent intervals.}
\label{tab:A3}
\end{table}

\begin{table}[!htb]\centering
\footnotesize\setlength{\tabcolsep}{3pt}\renewcommand{\arraystretch}{0.88}
\begin{tabular}{@{}l>{\raggedright\arraybackslash}p{0.3\columnwidth}>{\raggedright\arraybackslash}p{0.3\columnwidth}@{}}
\toprule
Model & All defects caught & False alarms \\
\midrule
Qwen3.5 9B & $-$0.053\newline \mbox{\scriptsize [$-$0.133, 0.024]} & $-$0.104\newline \mbox{\scriptsize [$-$0.158, $-$0.051]} \\
Qwen3.5 27B & +0.027\newline \mbox{\scriptsize [$-$0.026, 0.079]} & +0.059\newline \mbox{\scriptsize [0.008, 0.113]} \\
Haiku 5.5 & $-$0.040\newline \mbox{\scriptsize [$-$0.084, 0.004]} & $-$0.005\newline \mbox{\scriptsize [$-$0.024, 0.013]} \\
\bottomrule
\end{tabular}
\caption{The single agent without the instruction minus the team with it, one call of each arm on the 100 paired memos, for the three models below ceiling, with 95 percent intervals.}
\label{tab:A4}
\end{table}

\section{Signal detection analysis}

From the pooled rates we compute the two measures of equal-variance signal detection theory \citep{stanislaw1999calculation}, whose assumption of equal variance a single operating point cannot test. With $z$ the inverse of the standard normal distribution function, discrimination is $d' = z(H) - z(F)$ and the threshold is $c = -(z(H) + z(F))/2$. A lower $c$ means a reviewer more ready to flag. The measures are not computed for Sonnet 5.5, whose single agent flags no clean cell. Table~\ref{tab:A5} gives the changes. Against the single agent, the team's $d'$ does not change significantly on any of the three models and its $c$ falls on all three. The pooled rates are therefore consistent with a lower threshold and unchanged discrimination, and they do not establish that account. The intervals on $d'$ are wide. On Haiku 5.5 the upper bound corresponds to roughly the whole of the team's gain in defects caught, so a real improvement of that size is not excluded. No change in $d'$ is significant within any defect class either, and the estimates rise in some classes and fall in others. On Qwen3.5 27B and Haiku 5.5 the scoped team's threshold rises back toward the single agent's.

\begin{table}[!htb]\centering
\footnotesize\setlength{\tabcolsep}{3pt}\renewcommand{\arraystretch}{0.88}
\begin{tabular}{@{}>{\raggedright\arraybackslash}p{0.26\columnwidth}>{\raggedright\arraybackslash}p{0.22\columnwidth}>{\raggedright\arraybackslash}p{0.22\columnwidth}>{\raggedright\arraybackslash}p{0.22\columnwidth}@{}}
\toprule
 & Qwen3.5 9B & Qwen3.5 27B & Haiku 5.5 \\
\midrule
$d'_{\mathrm{team}} - d'_{\mathrm{single}}$, all defects & +0.01\newline \mbox{\scriptsize [$-$0.15, 0.17]} & $-$0.09\newline \mbox{\scriptsize [$-$0.29, 0.12]} & $-$0.03\newline \mbox{\scriptsize [$-$0.33, 0.23]} \\
$d'_{\mathrm{team}} - d'_{\mathrm{single}}$, bare & +0.13\newline \mbox{\scriptsize [$-$0.13, 0.40]} & not computed & not computed \\
$d'_{\mathrm{team}} - d'_{\mathrm{single}}$, masked & $-$0.11\newline \mbox{\scriptsize [$-$0.37, 0.14]} & +0.04\newline \mbox{\scriptsize [$-$0.23, 0.30]} & +0.17\newline \mbox{\scriptsize [$-$0.19, 0.51]} \\
$d'_{\mathrm{team}} - d'_{\mathrm{single}}$, arithmetic & +0.07\newline \mbox{\scriptsize [$-$0.16, 0.31]} & $-$0.21\newline \mbox{\scriptsize [$-$0.44, 0.04]} & $-$0.25\newline \mbox{\scriptsize [$-$0.57, 0.05]} \\
$d'_{\mathrm{scoped}} - d'_{\mathrm{single}}$, all defects & +0.01\newline \mbox{\scriptsize [$-$0.14, 0.16]} & $-$0.06\newline \mbox{\scriptsize [$-$0.26, 0.14]} & +0.14\newline \mbox{\scriptsize [$-$0.20, 0.49]} \\
$d'_{\mathrm{scoped}} - d'_{\mathrm{team}}$, all defects & 0.00\newline \mbox{\scriptsize [$-$0.14, 0.15]} & +0.02\newline \mbox{\scriptsize [$-$0.14, 0.19]} & +0.17\newline \mbox{\scriptsize [$-$0.08, 0.45]} \\
$c_{\mathrm{team}} - c_{\mathrm{single}}$ & $-$0.33\newline \mbox{\scriptsize [$-$0.42, $-$0.24]} & $-$0.20\newline \mbox{\scriptsize [$-$0.30, $-$0.11]} & $-$0.24\newline \mbox{\scriptsize [$-$0.39, $-$0.10]} \\
$c_{\mathrm{scoped}} - c_{\mathrm{team}}$ & $-$0.06\newline \mbox{\scriptsize [$-$0.13, 0.02]} & +0.09\newline \mbox{\scriptsize [0.01, 0.16]} & +0.17\newline \mbox{\scriptsize [0.05, 0.31]} \\
\bottomrule
\end{tabular}
\caption{Changes in $d'$ and in $c$ under the equal-variance model, with 95 percent intervals, on the three models below ceiling. By class, a common false-alarm rate is used. A class caught in full by either arm is not computed.}
\label{tab:A5}
\end{table}

\section{Further tables}

Tables A6 to A10 give the outcomes for whole memos, the rates by requirement, the repeated calls of the single agent's prompt, the misfiled flags and the share of flags that would be true.

\raggedbottom\par\medskip\noindent\begin{minipage}{\columnwidth}\centering
\footnotesize\setlength{\tabcolsep}{3pt}\renewcommand{\arraystretch}{0.88}
\begin{tabular}{@{}ll>{\raggedright\arraybackslash}p{0.2\columnwidth}>{\raggedright\arraybackslash}p{0.26\columnwidth}@{}}
\toprule
Model & Arm & Clean memos flagged, of 21 & Defective memos fully caught, of 279 \\
\midrule
Qwen3.5 9B & Single agent & 10 & 94 \\
 & Team & 13 & 141 \\
 & Scoped team & 18 & 151 \\
 & Three draws, any & 16 & 196 \\
Qwen3.5 27B & Single agent & 3 & 157 \\
 & Team & 5 & 186 \\
 & Scoped team & 6 & 172 \\
 & Three draws, any & 16 & 251 \\
Haiku 5.5 & Single agent & 0 & 174 \\
 & Team & 1 & 205 \\
 & Scoped team & 1 & 197 \\
 & Three draws, any & 3 & 234 \\
Sonnet 5.5 & Single agent & 0 & 275 \\
 & Team & 0 & 273 \\
 & Scoped team & 0 & 275 \\
 & Three draws, any & 1 & 279 \\
\bottomrule
\end{tabular}
\captionof{table}{Whole memos, decide-once instruction. Of the 21 memos with no defect, the number with at least one flag. Of the 279 memos with a defect, the number with every defect caught. The majority rule is omitted. Twenty-one memos with no defect are too few to estimate a rate for such files.}
\label{tab:A6}
\end{minipage}\par\medskip

\raggedbottom\par\medskip\noindent\begin{minipage}{\columnwidth}\centering
\footnotesize\setlength{\tabcolsep}{3pt}\renewcommand{\arraystretch}{0.88}
\begin{tabular}{@{}llll@{}}
\toprule
Model & Requirement & $H$ & $F$ \\
\midrule
Qwen3.5 9B & R1 & 0.68, 0.84, 0.89 & 0.143, 0.251, 0.325 \\
 & R2 & 0.80, 0.67, 0.70 & 0.012, 0.106, 0.065 \\
 & R3 & 0.38, 0.65, 0.65 & 0.082, 0.175, 0.219 \\
 & R4 & 0.77, 0.80, 0.81 & 0.561, 0.529, 0.561 \\
 & R5 & 0.67, 0.96, 0.90 & 0.128, 0.346, 0.336 \\
 & R6 & 0.38, 0.58, 0.64 & 0.022, 0.082, 0.093 \\
Qwen3.5 27B & R1 & 0.58, 0.80, 0.68 & 0.025, 0.094, 0.034 \\
 & R2 & 0.93, 0.87, 0.97 & 0.006, 0.012, 0.000 \\
 & R3 & 0.73, 0.74, 0.72 & 0.022, 0.093, 0.066 \\
 & R4 & 0.65, 0.66, 0.60 & 0.176, 0.214, 0.225 \\
 & R5 & 0.90, 0.97, 0.94 & 0.024, 0.028, 0.038 \\
 & R6 & 0.93, 0.97, 0.96 & 0.027, 0.011, 0.011 \\
Haiku 5.5 & R1 & 0.66, 0.75, 0.73 & 0.000, 0.005, 0.000 \\
 & R2 & 0.99, 1.00, 1.00 & 0.018, 0.018, 0.012 \\
 & R3 & 0.80, 0.87, 0.78 & 0.033, 0.049, 0.011 \\
 & R4 & 0.70, 0.63, 0.65 & 0.027, 0.075, 0.059 \\
 & R5 & 0.80, 0.99, 0.97 & 0.005, 0.005, 0.000 \\
 & R6 & 0.94, 1.00, 1.00 & 0.000, 0.000, 0.000 \\
Sonnet 5.5 & R1 & 1.00, 0.99, 1.00 & 0.000, 0.049, 0.010 \\
 & R2 & 1.00, 1.00, 1.00 & 0.000, 0.006, 0.000 \\
 & R3 & 1.00, 1.00, 0.99 & 0.000, 0.011, 0.005 \\
 & R4 & 0.96, 0.96, 0.97 & 0.000, 0.011, 0.000 \\
 & R5 & 1.00, 1.00, 1.00 & 0.000, 0.000, 0.000 \\
 & R6 & 1.00, 1.00, 1.00 & 0.000, 0.000, 0.000 \\
\bottomrule
\end{tabular}
\captionof{table}{Share of defects caught, $H$, and false alarms, $F$, by requirement, decide-once instruction, 300 memos. Each entry gives single agent, team and scoped team in that order. The numbers of defective and clean cells are 97 and 203 for R1, 130 and 170 for R2, 117 and 183 for R3, 113 and 187 for R4, 89 and 211 for R5, and 118 and 182 for R6. The Qwen3.5 9B model flags requirement R4 on more than half of its clean cells in every arm. It treats a multiple quoted to one decimal place as wrong because it is rounded.}
\label{tab:A7}
\end{minipage}\par\medskip

\raggedbottom\par\medskip\noindent\begin{minipage}{\columnwidth}\centering
\footnotesize\setlength{\tabcolsep}{3pt}\renewcommand{\arraystretch}{0.88}
\begin{tabular}{@{}>{\raggedright\arraybackslash}p{0.22\columnwidth}>{\raggedright\arraybackslash}p{0.165\columnwidth}>{\raggedright\arraybackslash}p{0.165\columnwidth}>{\raggedright\arraybackslash}p{0.165\columnwidth}>{\raggedright\arraybackslash}p{0.165\columnwidth}@{}}
\toprule
 & Qwen3.5 9B & Qwen3.5 27B & Haiku 5.5 & Sonnet 5.5 \\
\midrule
$H^{\mathrm{masked}}$, each repeat & 0.424\newline 0.445\newline 0.432\newline 0.559\newline @Temp=0 & 0.821\newline 0.764\newline 0.738\newline 0.734\newline @Temp=0 & 0.790\newline 0.803\newline 0.769\newline 0.786 & 1.000\newline 1.000\newline 1.000\newline 1.000 \\
$H$, each repeat & 0.586\newline 0.592\newline 0.556\newline 0.613\newline @Temp=0 & 0.854\newline 0.800\newline 0.792\newline 0.792\newline @Temp=0 & 0.825\newline 0.830\newline 0.798\newline 0.830 & 0.994\newline 0.995\newline 0.994\newline 0.998 \\
$F$, each repeat & 0.191\newline 0.185\newline 0.172\newline 0.160\newline @Temp=0 & 0.070\newline 0.078\newline 0.085\newline 0.047\newline @Temp=0 & 0.013\newline 0.011\newline 0.013\newline 0.008 & 0.000\newline 0.000\newline 0.000\newline 0.001 \\
Team: $H^{\mathrm{masked}}$, $H$, $F$ & 0.642\newline 0.733\newline 0.254 & 0.817\newline 0.834\newline 0.076 & 0.891\newline 0.877\newline 0.025 & 0.996\newline 0.991\newline 0.013 \\
\bottomrule
\end{tabular}
\captionof{table}{Repeated calls of the single agent's prompt, decide-once instruction, 300 memos. For the Claude models the four repeats are the single call and the three draws, in that order. For the Qwen3.5 models the three draws at temperature 0.7 are shown, with the single call at temperature 0 beside them, marked @Temp=0. On Haiku 5.5 the team's masked rate and its false-alarm rate both lie outside the range of the four repeats. No repeat of the team or the scoped team exists.}
\label{tab:A8}
\end{minipage}\par\medskip

\raggedbottom\par\medskip\noindent\begin{minipage}{\columnwidth}\centering
\footnotesize\setlength{\tabcolsep}{3pt}\renewcommand{\arraystretch}{0.88}
\begin{tabular}{@{}l>{\raggedright\arraybackslash}p{0.165\columnwidth}>{\raggedright\arraybackslash}p{0.165\columnwidth}>{\raggedright\arraybackslash}p{0.165\columnwidth}>{\raggedright\arraybackslash}p{0.19\columnwidth}@{}}
\toprule
Model & Single agent, flags and misfiled & Team & Scoped team & $\Delta F$, misfiled set aside \\
\midrule
Qwen3.5 9B & 182 and 16 & 288 and 28 & 310 and 25 & +0.083\newline \mbox{\scriptsize [0.053, 0.112]} \\
Qwen3.5 27B & 53 and 5 & 86 and 26 & 71 and 14 & +0.011\newline \mbox{\scriptsize [$-$0.006, 0.027]} \\
Haiku 5.5 & 15 and 1 & 28 and 17 & 15 and 2 & $-$0.003\newline \mbox{\scriptsize [$-$0.011, 0.005]} \\
Sonnet 5.5 & 0 and 0 & 15 and 8 & 3 and 2 & +0.006\newline \mbox{\scriptsize [0.002, 0.011]} \\
\bottomrule
\end{tabular}
\captionof{table}{Flags on clean cells and, of those, misfiled flags, whose quoted sentence is the defective sentence of a defect under another requirement in the same memo. Decide-once instruction, 1,136 clean cells. The last column is $\Delta F$ with misfiled flags set aside, with a 95 percent interval. Sentences are matched by text after removing case and punctuation. A looser definition that also counts the correct statement of a fact whose other statement is wrong gives the same counts on Qwen3.5 27B and the two Claude models, and 24, 40 and 43 on Qwen3.5 9B.}
\label{tab:A9}
\end{minipage}\par\medskip

\raggedbottom\par\medskip\noindent\begin{minipage}{\columnwidth}\centering
\footnotesize\setlength{\tabcolsep}{3pt}\renewcommand{\arraystretch}{0.88}
\begin{tabular}{@{}>{\raggedright\arraybackslash}p{0.22\columnwidth}>{\raggedright\arraybackslash}p{0.165\columnwidth}>{\raggedright\arraybackslash}p{0.165\columnwidth}>{\raggedright\arraybackslash}p{0.165\columnwidth}>{\raggedright\arraybackslash}p{0.165\columnwidth}@{}}
\toprule
 & Qwen3.5 9B & Qwen3.5 27B & Haiku 5.5 & Sonnet 5.5 \\
\midrule
Single agent & 0.69\newline 0.17\newline 0.07 & 0.91\newline 0.47\newline 0.26 & 0.97\newline 0.77\newline 0.56 & 1.00\newline 1.00\newline 1.00 \\
Team & 0.63\newline 0.13\newline 0.06 & 0.87\newline 0.37\newline 0.18 & 0.95\newline 0.65\newline 0.42 & 0.98\newline 0.80\newline 0.60 \\
Scoped team & 0.62\newline 0.13\newline 0.05 & 0.88\newline 0.41\newline 0.21 & 0.97\newline 0.77\newline 0.57 & 1.00\newline 0.95\newline 0.88 \\
Three draws, majority & 0.71\newline 0.18\newline 0.08 & 0.91\newline 0.48\newline 0.27 & 0.99\newline 0.94\newline 0.87 & 1.00\newline 1.00\newline 1.00 \\
Three draws, any & 0.59\newline 0.11\newline 0.05 & 0.76\newline 0.22\newline 0.10 & 0.95\newline 0.63\newline 0.39 & 1.00\newline 0.98\newline 0.96 \\
\bottomrule
\end{tabular}
\captionof{table}{Share of flags that would be true at three values of the defect share $p$, from the unrounded pooled rates behind Table~\ref{tab:2} and assuming they carry over. Each entry gives $p$ of 0.37, 0.05 and 0.02, from top to bottom.}
\label{tab:A10}
\end{minipage}\par\medskip

\end{document}